\pdfoutput=1
\documentclass[graybox]{svmult}

\usepackage{mathptmx}       % selects Times Roman as basic font
\usepackage{helvet}         % selects Helvetica as sans-serif font
\usepackage{courier}        % selects Courier as typewriter font
\usepackage{type1cm}        % activate if the above 3 fonts are
\usepackage{makeidx}         % allows index generation
\usepackage{graphicx}        % standard LaTeX graphics tool
\usepackage{multicol}        % used for the two-column index
\usepackage[bottom]{footmisc}% places footnotes at page bottom

\usepackage{multirow}
\usepackage{multicol}
\usepackage{subfigure} %In order to put two tables/figures side by side and with subcaptions
\usepackage{dcolumn}
\usepackage[nomarkers,notablist,nofiglist]{endfloat}
\usepackage{longtable}
\usepackage{newclude}

\makeindex             % used for the subject index
\def \vY{\mbox{\boldmath $Y$ \unboldmath}\!\!}
\def \vw{\mbox{\boldmath $w$ \unboldmath}\!\!}

\def \vbeta{\mbox{\boldmath $\beta$ \unboldmath}\!\!}
\def \vtheta{\mbox{\boldmath $\theta$ \unboldmath}\!\!}
\def \vrho{\mbox{\boldmath $\rho$ \unboldmath}\!\!}
\def \vepsilon{\mbox{\boldmath $\epsilon$ \unboldmath}\!\!}

\begin{document}

\title*{High-Dimensional Classification for Brain Decoding}

\titlerunning{High-Dimensional Brain Decoding}
\author{Nicole Croteau, Farouk S. Nathoo, Jiguo Cao, Ryan Budney}
\institute{Corresponding Author: Farouk S. Nathoo, Department of Mathematics and Statistics, University of Victoria, \email{nathoo@uvic.ca}}
\maketitle

\abstract{Brain decoding involves the determination of a subject's cognitive state or an associated stimulus from functional neuroimaging data measuring brain activity. In this setting the cognitive state is typically characterized by an element of a finite set, and the neuroimaging data comprise voluminous amounts of spatiotemporal data measuring some aspect of the neural signal. The associated statistical problem is one of classification from high-dimensional data. We explore the use of functional principal component analysis, mutual information networks, and persistent homology for examining the data through exploratory analysis and for constructing features characterizing the neural signal for brain decoding. We review each approach from this perspective, and we incorporate the features into a classifier based on symmetric multinomial logistic regression with elastic net regularization. The approaches are illustrated in an application where the task is to infer, from brain activity measured with magnetoencephalography (MEG), the type of video stimulus shown to a subject.}

\section{Introduction}
\label{sec:1}

Recent advances in techniques for measuring brain activity through neuroimaging modalities such as functional magnetic resonance imaging (fMRI), electroencephalography (EEG), and magnetoencephalography (MEG) have demonstrated the possibility of decoding a person's conscious experience based only on non-invasive measurements
of their brain signals (Hayens and Reese, 2006). Doing so involves uncovering the relationship between the recorded signals and the conscious experience that may then provide insight into the underlying mental process. Such decoding tasks arise in a number of areas, for example, the area of brain-computer interfaces, where humans can be trained to use their brain activity to control artificial devices. At the heart of this task is a classification problem where the neuroimaging data comprise voluminous amounts of spatiotemporal observations measuring some aspect of the neural signal across an array of sensors outside the head (EEG, MEG) or voxels within the brain (fMRI). With neuroimaging data the classification problem can be challenging as the recorded brain signals have a low signal-to-noise ratio and the size of the data leads to a high-dimensional problem where it is easy to overfit models to data when training a classifier. Overfitting will impact negatively on the degree of generalization to new data and thus must be avoided in order for solutions to be useful for practical application. 

Neuroimaging classification problems have been studied extensively in recent years primarily in efforts to develop biomarkers for neurodegenerative diseases and other brain disorders. A variety of techniques have been applied in this context, including support vector machines (Chapelle et al., 1999), Gaussian process classification (Rasmussen, 2004), regularized logistic regression (Tomioka et al. 2009), and neural networks (Ripely, 1994; Neal and Zhang, 2006).  Decoding of brain images using Bayesian approaches is discussed by Friston et. al (2008). While a variety of individual classifiers or an ensemble of classifiers may be applied in any given application, the development of general approaches to constructing features that successfully characterize the signal in functional neuroimaging data is a key open problem. In this article we explore the use of some recent approaches developed in statistics and computational topology as potential solutions to this problem. More specifically, we consider how the combination of functional principal component analysis (Ramsay and Silverman, 2005), persistent homology (Carlson, 2009), and network measures of brain connectivity (Rubinov and Sporns, 2010) can be used to (i) explore large datasets of recorded brain activity and (ii) construct features for the brain decoding problem. 

The objectives of this article are threefold. First, we wish to introduce the brain decoding problem to researchers working in the area of high-dimensional data analysis. This challenging problem serves as a rich arena for applying recent advances in methodology. Moreover, the specific challenges associated with the brain decoding problem (e.g. low signal-to-noise ratio; spatiotemporal data) can help to further motivate the development of new methods. Our second objective is to describe how functional principal component analysis (FPCA), persistent homology, and network measures of brain connectivity can be used to explore such data and construct features. To our knowledge, FPCA and persistent homology have not been previously considered as approaches for constructing features for brain decoding. 

Our third and final objective is to illustrate these methods in a real application involving MEG data, where the goal is to explore variability in the brain data and to use the data to infer the type of video stimulus shown to a subject from a 1-second recording obtained from 204 MEG sensors with the signal at each channel sampled at a frequency of 200Hz. Each sample thus yields $204 \times 200 = 40800$ observations of magnetic field measurements outside the head. The goal is to decode which of five possible video stimuli was shown to the subject during the recording from these measurements. The data arising from a single sample are shown in Figure 1, where panel (a) depicts the brain signals recorded across all sensors during the 1-second recording, and panel (b) depicts the variance of the signal at each location. From panel (b) we see that in this particular sample the stimulus evoked activity in the regions associated with the temporal and occipital lobes of the brain. The entire dataset for the application includes a total of 1380 such samples (727 training; 653 test) obtained from the same subject which together yield a dataset of roughly 6 gigabytes in compressed format.

Functional principal component analysis (FPCA) is the extension of standard finite-dimensional PCA to the setting where the response variables are functions, a setting referred to as functional data. For clarity, we note here that the use of the word 'functional' in this context refers to functional data as just described, and is not to be confused with functional neuroimaging data which refers to imaging data measuring the function of the brain. Given a sample of functional observations (e.g. brain signals) with each signal assumed a realization of a square-integrable stochastic process over a finite interval, FPCA involves the estimation of a set of eigenvalue-eigenfunction pairs that describe the major vibrational components in the data. These components can be used to define features for classification through the projection of each signal onto a set of estimated eigenfunctions characterizing most of the variability in the data. This approach has been used recently for the classification of genetic data by Leng and M\"{u}ller (2005) who use FPCA in combination with logistic regression to develop a classifier for temporal gene expression data. 

An alternative approach for exploring the patterns in brain signals is based on viewing each signal obtained at a voxel or sensor as a point in high-dimensional Euclidean space. The collection of signals across the brain then forms a point cloud in this space, and the shape of this point cloud can be described using tools from topological data analysis (Carlson, 2009). In this setting the data are assumed clustered around a familiar object like a manifold, algebraic variety or cell complex and the objective is to describe (estimate some aspect of) the topology of this object from the data. The subject of persistent homology can be seen as a concrete manifestation of this idea, and provides a novel method to discover non-linear features in data.  With the same advances in modern computing technology that allow for the storage of large datasets, persistent homology and its variants can be implemented. Features derived from persistent homology have recently been found useful for classification of hepatic lesions (Adcock et al., 2014) and persistent homology has been applied for the analysis of structural brain images (Chung et al., 2009; Pachauri et al., 2011). Outside the arena of medical applications, Sethares and Budney (2013) use persistent homology to study topological structures in musical data. Recent work in Heo et al. (2012) connects computational topology with the traditional analysis of variance and combines these approaches for the analysis of multivariate orthodontic landmark data derived from the maxillary complex. The use of persistent homology for exploring structure of spatiotemporal functional neuroimaging data does not appear to have been considered previously. 

Another alternative for exploring patterns in the data is based on estimating and summarizing the topology of an underlying network. Networks are commonly used to explore patterns in both functional and structural neuroimaging data. With the former, the nodes of the network correspond to the locations of sensors/voxels and the links between nodes reflect some measure of dependence between the time series collected at pairs of locations. To characterize dependence between time series, the mutual information, a measure of shared information between two time series is a useful quantity as it measures both linear and nonlinear dependence (Zhou et al., 2009), the latter being potentially important when characterizing dependence between brain signals (Stam et al., 2003). Given such a network, the corresponding topology can be summarized with a small number of meaningful measures such as those representing the degree of small-world organization (Rubinov and Sporns, 2010). These measures can then be explored to detect differences in the network structure of brain activity across differing stimuli and can be further used as features for brain decoding. 

The remainder of the paper is structured as follows. Section 2 describes the classifier and discusses important considerations for defining features. Section 3 provides a review of FPCA from the perspective of exploring functional neuroimaging data. Sections 4 and 5 discuss persistent homology and mutual information networks, respectively, as approaches for characterizing the interaction of brain signals and defining nonlinear features for classification. Section 6 presents an application to the decoding of visual stimuli from MEG data, and Section 7 concludes with a brief discussion.

\section{Decoding Cognitive States from Neuroimaging Data}
\label{sec:2}

Let us assume we have observed functional neuroimaging data $\vY=\{y_{i}(t), i=1,\dots,n;\, t=1,\dots,T\}$ where $y_{i}(t)$ denotes the signal of brain activity measured at the $i^{th}$ sensor or voxel. We assume that there is a well-defined but unknown cognitive state corresponding to these data that can be represented by the label $C \in \{1,\dots,K\}$. The decoding problem is that of recovering $C$ from $\vY$. A solution to this problem involves first summarizing $\vY$ through an $m$-dimensional vector of features 
$\vY_{f} = (Y_{f_{1}},\dots,Y_{f_{m}})'$  and then applying a classification rule $R^{m} \rightarrow \{1,\dots,K\}$ to obtain the predicted state. A solution must specify how to construct the features and define the classification rule, and we assume there exists a set of training samples $\vY_{l} = \{y_{li}(t), i=1,\dots,n;\, t=1,\dots,T\},\, l=1,\dots,L$ with \emph{known} labels $C_{l}, \, l=1,\dots, L$ for doing this.

To define the classification rule we model the training labels with a multinomial distribution where the class probabilities are related to features through a symmetric multinomial logistic regression (Friedman et al., 2010) having form
\begin{equation}
\label{logit}
Pr(C = j) = \frac{\exp(\beta_{0j} + \vbeta_{j}'\vY_{f})}{\sum_{k=1}^{K}   \exp(\beta_{0k} + \vbeta_{k}'\vY_{f})} ,\,\,\, j=1,\dots,K
\end{equation}
with parameters $\vtheta = (\beta_{01},\vbeta_{1}',\dots,\beta_{0K},\vbeta_{K}')'$. As the dimension of the feature vector will be large relative to the number of training samples we estimate $\vtheta$ from the training data using regularized maximum likelihood. This involves maximizing a penalized log-likelihood where the likelihood is defined by the symmetric multinomial logistic regression and we incorporate an elastic net penalty (Zou and Hastie, 2006). Optimization is carried using cyclical coordinate descent as implemented in the \emph{glmnet} package (Friedman et al., 2010) in R. The two tuning parameters weighting the $l_{1}$ and $l_{2}$ components of the elastic net penalty are chosen using cross-validation over a grid of possible values. Given $\hat{\vtheta}$ the classification of a new sample with unknown label is based on computing the estimated class probabilities from (\ref{logit}) and choosing the state with the highest estimated value. 

To define the feature vector $\vY_{f}$ from $\vY$ we consider two aspects of the neural signal that are likely important for discriminating cognitive states. The first aspect involves the shape and power of the signal at each location. These are local features computed at each voxel or sensor irrespective of the signal observed at other locations. The variance of the signal computed over all time points is one such feature that will often be useful for discriminating states, as different states may correspond to different locations of activation, and these locations will have higher variability in the signal. The second aspect is the functional connectivity representing how signals at different locations interact. Rather than being location specific, such features are global and may help to resolve the cognitive state in the case where states correspond to differing patterns of interdependence among the signals across the brain. From this perspective we next briefly describe FPCA, persistent homology, and mutual information networks as approaches for exploring these aspects of functional neuroimaging data, and further how these approaches can be used to define features for classification.

\section{Functional Principal Component Analysis}
\label{sec:3}

Let us fix a particular location $i$ of the brain or sensor array. At this specific location we observe a sample of curves $y_{li}(t)$, $l=1,\dots,L$ where the size of the sample corresponds to that of the training set. We assume that each curve is an independent realization of a square-integrable stochastic process $y_{i}(t)$ on $[0,T]$ with mean $E[y_{i}(t)] = \mu_{i}(t)$ and covariance $Cov[y_{i}(t),y_{i}(s)] = G_{i}(s,t)$. The process can be written in terms of the Karhunen-Lo\`eve representation (Leng and M\"{u}ller, 2005)
\begin{equation}
\label{KL}
y_{i}(t) = \mu_{i}(t) + \sum_{m}\epsilon_{mi}\rho_{mi}(t)
\end{equation}
where $\{\rho_{mi}(t)\}$ is a set of orthogonal functions referred to as the functional principal components (FPCs) with corresponding coefficients 
\begin{equation}
\label{KL_coeff}
\epsilon_{mi} = \int_{0}^{T}(y_{i}(t)-\mu_{i}(t))\rho_{mi}(t)dt
\end{equation}
with $E[\epsilon_{mi}] = 0$, $Var[\epsilon_{mi}] = \lambda_{mi}$ and the variances are ordered so that $\lambda_{1i} \ge \lambda_{2i} \ge \cdots$ with  $\sum_{m}\lambda_{mi} < \infty$. The total variability of process realizations about $\mu_{i}(t)$ is governed by the random coefficients $\epsilon_{mi}$ and in particular by the corresponding variance $\lambda_{mi}$, with relatively higher values corresponding to FPCs that contribute more to this total variability. 

Given the $L$ sample realizations, the estimates of $\mu_{i}(t)$ and of the first few FPCs can be used to explore the dominant modes of variability in the observed brain signals at location $i$. The mean curve is estimated simply as $\hat{\mu}_{i}(t) = \frac{1}{L}\sum_{l=1}^{L}y_{li}(t)$ and from this the covariance function $G_{i}(s,t)$ is estimated $\hat{G_{i}} = \hat{Cov}[y_{i}(s_{k}), y_{i}(s_{l})]$ using the empirical covariance over a grid of points $s_{1},\dots,s_{S} \in [0,T]$. The FPCs are then estimated through the spectral decomposition of $\hat{G}_{i}$ (see e.g. Ramsay and Silverman, 2005) with the eigenvectors yielding the estimated FPCs evaluated at the grid points, $\hat{\vrho}_{mi} = (\hat{\vrho}_{mi}(s_{1}),\dots, \hat{\vrho}_{mi}(s_{S}))'$, and the corresponding eigenvalues being the estimated variances $\hat{\lambda}_{mi}$ for the coefficients $\epsilon_{mi}$ in (\ref{KL}). The fraction of the sample variability explained by the first $M$ estimated FPCs can then be expressed as $FVE(M) = \sum_{m=1}^{M}\hat{\lambda}_{mi}/\sum_{m}\hat{\lambda}_{mi}$ and this can be used to choose a nonnegative integer $M_{i}$ so that the predicted curves
$$
\hat{y}_{li}(t) = \hat{\mu}_{i}(t) + \sum_{m=1}^{M_{i}}\hat{\epsilon}_{lmi}\hat{\rho}_{mi}(t)
$$
explain a specified fraction of the total sample variability. We note that in producing the predicted curve a separate realization of the coefficients $\epsilon_{mi}$ from (\ref{KL}) is estimated from each observed signal using (\ref{KL_coeff}) and, for a given $m$, the estimated coefficients $\hat{\vepsilon}_{mi} = \{\hat{\epsilon}_{lmi},\, l=1,\dots,L\}$ are referred to as the order-$m$ FPC scores which represent between subject variability in the particular mode of variation represented by $\hat{\rho}_{mi}(t)$. The scores are thus potentially useful as features for classification. 

We compute the FPC scores $\hat{\vepsilon}_{mi},\, m=1,\dots,M_{i}$ separately at each location $i=1,\dots,n$. For a given location the number of FPCs, $M_{i}$, is chosen to be the smallest integer such that the $FVE(M_{i}) \ge 0.9$. Thus the number of FPCs, $M_{i}$, will vary across locations but typically only a small number will be required. Locations requiring a relatively greater number of FPCs will likely correspond to locations where the signal is more complex. The total number of features introduced by our application of FPCA for brain decoding is then $\sum_{i=1}^{n}M_{i}$. The FPCs and the associated FPC scores are computed using the \emph{fda} package in R (Ramsay and Silverman, 2005).

\section{Persistent Homology}
\label{sec:4}

Let us now fix a particular sample $l$ from the training set and consider the collection of signals, $y_{li}(t)$, observed over all locations $i=1,\dots,n$ for that sample. Each signal is observed over the same set of $T$ equally-spaced time points $\vY_{li} = (y_{li}(1),\dots,y_{li}(T))'$ and can thus be considered a point in $R^{T}$. The sample of signals across the brain/sensors then forms a point cloud in $R^{T}$. For example, the single sample depicted in Figure 1, panel (a), represents a cloud of $n=204$ points in $R^{200}$. Using tools from topological data analysis we aim to identify topological structures associated with this point cloud and to use such structures as features for brain decoding. 

As a metric inducing the topology we require a measure of statistical dependence that will collate both correlated and anti-correlated signals. We therefore employ the absolute Pearson correlation distance metric $D(\vY_{li},\vY_{lj}) = 1 - \rho(\vY_{li},\vY_{lj})^{2}$ where $\rho(\vY_{li},\vY_{lj})$ is the sample correlation between signals at locations $i$ and $j$. We focus specifically on persistent homology, which attempts to identify the connected components, loops, and voids of an associated manifold that we assume the point cloud has been sampled from. The general idea is to approximate the manifold using a simpler object, a simplicial complex, for which the homology (a characterization of the holes in the shape) may be calculated. A sequence of such approximations covering a range of scales is considered and the features that persist over a large range are considered as intrinsic to the data. We provide here only a general description that emphasizes basic concepts and intuition for the construction of features for brain decoding. A more detailed but still gentle introduction to persistent homology including the required definitions and results from simplicial homology theory and group theory is provided by Zhu (2013).

Given the point cloud of $n$ signals and the metric based on correlation distance, we consider a covering of the points by balls of radius $\epsilon$, and for any such covering, we associate a simplicial complex for which the homology classes can be calculated. The $p$-th Betti number, $\textrm{Betti}_{p}$,  can be thought of as representing the number of p-dimensional holes, which for $p=0,1,2$ corresponds to the \emph{number of connected components}, \emph{loops} and \emph{voids}, respectively. The value of $\epsilon$ is then varied over many possible values creating a filtration (an increasing sequence of simplicial complexes). The growing radius $\epsilon$ corresponds to allowing signals with smaller values of the squared-correlation to be linked together to form simplices in the simplicial complex. The homology classes are calculated at each value of $\epsilon$ to determine how the system of holes changes. Persistent features remain over a relatively large range of $\epsilon$ values and are thus considered as signal in the data. 

The nature of change with respect to each dimension $p$ can be depicted graphically using a barcode plot, a plot that tracks the birth and death of holes as $\epsilon$ varies. Features in the barcode that are born and then quickly die are considered topological noise, while features that persist are considered indicative of signal in the underlying topology. If the barcode plot of dimension $p=0$ reveals signal and the higher-dimensional barcodes do not, the data are clustered around a metric tree. If both the $p=0$ and $p=1$ barcodes reveal signal and the $p=2$ barcode plot does not, the data are clustered around a metric graph.  A metric graph is indicative of multiple pathways for signals to get between two sensors/voxels. For barcodes of dimension $p>1$ the details can be rather subtle to sort through. For the sample considered in Figure 1, panel (a), the barcodes for each dimension $p=0, 1, 2$ are depicted in the first column of Figure 2. For a given barcode plot, the Betti number for fixed $\epsilon$ is computed as the number of bars above it. For $p=0$ (Figure 2, first row and first column), $\textrm{Betti}_{0} = 204$ connected components are born at $\epsilon=0$ corresponding to each of the MEG sensors. $\textrm{Betti}_{0}$ decreases rapidly as $\epsilon$ increases and it appears that between two to four connected components persist over a wide range of $\epsilon$ values. The barcode plot for dimension $p=1$ (Figure 2, second row and first column) also appears to have features that are somewhat significant, but the $p=2$ barcodes are relatively short, indicating noise. Taken together this can be interpreted as there being many loops in the point cloud. The data resemble a metric graph with some noise added. An equivalent way to depict the persistence of features is through a persistence diagram, which is a scatter plot comparing the birth and death $\epsilon$ values for each hole. The persistence diagrams corresponding to each barcode are are depicted in the second column of Figure 2.

As for interpretation in the context of functional neuroimaging data, the number of connected components ($\textrm{Betti}_{0}$) represents a measure of the overall connectivity 
or synchronization between sensors, with smaller values of $\textrm{Betti}_{0}$ corresponding to a greater degree of overall synchrony.  We suspect that the number of loops ($\textrm{Betti}_{1}$) corresponds to the density of 'information pathways' with higher values corresponding to more complex structure having more pathways. The number of voids ($\textrm{Betti}_{2}$) may be related to the degree of segregation of the connections. If a void was to persist through many values of $\epsilon$, then we may have a collection of locations/sensors that are not communicating. Thus the larger the value of $\textrm{Betti}_{2}$, the more of these non-communicative spaces there may be.

For each value of $p$, $p=0,1,2$, we construct features for classification by extracting information from the corresponding barcode by considering the persistence of each hole appearing at some point in the barcode. This is defined as the difference between the corresponding death and birth $\epsilon$ values for a given hole. This yields a sample of persistence values for each barcode. Summary statistics computed from this sample are then used as features for classification. In particular, we compute the total persistence, $PM_{p}$, which is defined as one-half of the sum of all persistence values,  and we also compute the variance, skewness, and kurtosis of the sample leading to additional features denoted as $PV_{p}$, $PS_{p}$, $PK_{p}$, respectively. In total we obtain 12 global features from persistent homology.

\section{Mutual Information Networks}
\label{sec:5}

Let us again fix a particular sample $l$ from the training set and consider the collection of signals, $y_{li}(t)$, observed over all locations $i=1,\dots,n$ for the given sample. For the moment we will suppress dependence on training sample $l$ and let $\vY_{i} = (Y_{i}(1),\dots,Y_{i}(T))'$ denote the time series recorded at location $i$. We next consider a graph theoretic approach that aims to characterize the global connectivity in the brain with a small number of neurobiologically meaningful measures. This is achieved by estimating a weighted network from the time series where the sensors/voxels correspond to the nodes of the network and the links $\hat{\vw}=(\hat{w}_{ij})$ represent the connectivity, where $\hat{w}_{ij}$ is a measure of statistical dependence estimated from $\vY_{i}$ and $\vY_{j}$. 

As a measure of dependence we consider the mutual information which quantifies the shared information between two time series and measures both linear and nonlinear dependence. The coherence between $\vY_{i}$ and $\vY_{j}$ at frequency $\lambda$ is a measure of correlation in frequency and is defined as $coh_{ij}(\lambda) = |f_{ij}(\lambda)|^{2}/(f_{i}(\lambda)*f_{j}(\lambda))$ where $f_{ij}(\lambda)$ is the cross-spectral density between $\vY_{i}$ and $\vY_{j}$ and $f_{i}(\lambda)$, $f_{j}(\lambda)$ are the corresponding spectral densities for each process (see e.g. Shumway and Stoffer, 2011). The mutual information within frequency band $[\lambda_{1},\lambda_{2}]$ is then 
$$
\delta_{ij} = -\frac{1}{2\pi} \int_{\lambda_{1}}^{\lambda_{2}}\log(1-coh_{ij}(\lambda))d\lambda
$$
and the network weights are defined as $w_{ij} = \sqrt{1-\exp(-2\delta_{ij})}$ which gives a measure of dependence lying in the unit interval (Joe, 1989). The estimates $\hat{\vw}$ are based on values $\lambda_{1}=0$, $\lambda_{2}=0.5$, and computed using the MATLAB toolbox for functional connectivity (Zhou et al., 2009). After computing the estimates of the network matrices we retained only the top 20\% strongest connections and set the remaining weights to $\hat{w}_{ij} = 0$. 

We summarize the topology of the network obtained from each sample with seven graph-theoretic measures, each of which can be expressed explicitly as a function of $\hat{\vw}$ (see e.g. Rubinov and Sporns, 2010). In computing the measures, the distance between any two nodes is taken as $\hat{w}_{ij}^{-1}$:
\begin{enumerate}
\item Characteristic path length: the average shortest path between all pairs of nodes.
\item Global efficiency: the average inverse shortest path length between all pairs of nodes.
\item Local efficiency: global efficiency computed over node neighbourhoods.
\item Clustering coefficient: an average measure of the prevalence of clustered connectivity around individual nodes.
\item Transitivity: a robust variant of the clustering coefficient.
\item Modularity: degree to which the network may be subdivided into clearly delineated and non-overlapping groups.
\item Assortativity coefficient: correlation coefficient between the degrees of all nodes on two opposite ends of a link.
\end{enumerate}
The seven measures are computed for each training sample and used as global features for brain decoding. 

\section{Example Application: Brain Decoding from MEG}
\label{sec:6}

In 2011 the International Conference on Artificial Neural Networks (ICANN) held an MEG mind reading contest sponsored by the PASCAL2 Challenge Programme. The challenge task was to infer from brain activity, measured with MEG, the type of a video stimulus shown to a subject. The experimental paradigm involved one male subject who watched alternating video clips from five video types while MEG signals were recorded at $n=204$ sensor channels covering the scalp. The different video types are: 
\begin{enumerate}
\item Artificial: screen savers showing animated shapes or text.
\item Nature: clips from nature documentaries, showing natural scenery like mountains or oceans.
\item Football: clips taken from (European) football matches of Spanish La Liga.
\item Mr. Bean: clips from the episode �\emph{Mind the Baby, Mr. Bean}� of the Mr. Bean television series.
\item Chaplin: clips from the �\emph{Modern Times}� feature film, starring Charlie Chaplin.
\end{enumerate}
The experiment involved two separate recording sessions that took place on consecutive days. The organizers released a series of 1-second MEG recordings in random order which were downsampled to 200Hz. A single recording is depicted in Figure 1, and the data comprise a total of 1380 such recordings. Of these, 677 recordings are labelled training samples from the first day of the experiment and 653 are unlabelled test samples from the second day of the experiment. Thus aside from the challenge of decoding the stimulus associated with test samples an additional challenge arises in that the training and test sets are from different days, leading to a potential domain shift problem. To aid contestants with this problem the organizers released a small additional set of 50 labelled training samples from day two. The objective was to use the 727 labelled training samples to build a classifier, and the submissions were judged based on the overall accuracy rate for decoding the stimulus of the test samples. The overall winning team obtained an accuracy rate of 68.0\%, which was followed by 63.2\% for the second place entry, and the remaining scores ranged from  62.8\% - 24.2\%. Full details of the competition and results are available in Klami et al. (2011). Following the competition, the labels for the 653 test samples were also released. Our objective is to apply the techniques described in this article to the ICANN MEG dataset and to compare the resulting decoding accuracy rates to those obtained in the actual competition. All rules of the competition were followed and the test data were only used to evaluate our approach, as in the competition. 

Examination of the training data reveals the detrended variance of the signal at each sensor to be an important feature for discriminating the stimuli. This is as expected (see discussion in Section 2) and so all classifiers we consider include this feature. Experimentation (using only the training data) with classifiers excluding the detrended variance indicated that this is by far the most important feature and the predicted accuracy rates we obtain from cross-validation drop significantly when this feature is excluded. In Figure 3 we illustrate the average spatial variation of this feature for each of the five stimuli. Differing patterns are seen for each class. For example, in the 'Chaplin' class, the signal exhibits greatest power in sensors representing the occipital and parietal lobes; whereas, for the 'Football' class we see the greatest power in sensors representing the left and right frontal lobes. Including the detrended variance at each sensor yields 204 features to be added to the classifier.

To derive additional features we applied FPCA to all of the training samples separately at each sensor. Figure 4 shows the first three functional principal components. The first FPC, depicted in panel (a), seems to capture the overall level of the signal. The second FPC, depicted in panel (b), appears to represent an overall trend and the third FPC, depicted in panel (c), is a mode of variation having a 'U' or an inverted 'U' shape. At each sensor, we included as features the minimum number of FPC scores required to explain 90\% of the variability at that sensor across the training samples. The distribution of the number of scores used at each sensor is depicted in Figure 4, panel (d). At most sensors either $M_{i}=2$ or $M_{i}=3$ FPC scores are used as features, and overall, FPCA introduces 452 features. The spatial variability of the first FPC score is depicted in Figure 5. Differing spatial patterns across all of the stimuli are visible, in particular for the 'Chaplin' class, which tends to have elevated first FPC scores at many sensors. 

Persistent homology barcodes of dimension $p=0,1,2$ were computed using the \emph{TDA} package in R (Fasy et al., 2014) for all training samples and the 12 summary features $PM_{p}, PV_{p}, PS_{p}, PK_{p}$, $p=0,1,2$ were extracted from the barcodes. To determine the potential usefulness for classification we compared the mean of each of these features across the five stimuli classes using one-way analysis of variance. In most cases the p-value corresponding to the null hypothesis of equality of the mean across all groups was less than 0.05, with the exception of $PK_{0}$ (p-value = .36) and $PM_{1}$ (p-value = .34). Mutual information weighted networks were also computed for each training sample and the seven graph theory measures discussed in Section 5 were calculated. Analysis of variance comparing the mean of each graph measure across stimuli classes resulted in p-values less than 0.001 for all seven features. This initial analysis indicates that both types of features, particularly the network features, may be useful for discriminating the stimuli for these data.

We considered a total of seven classifiers based on the setup described in Section 2 each differing with respect to the features included. The features included in each of the classifiers are indicated in Table 1. The simplest classifier included only the detrended variance (204 features) and the most complex classifier included the detrended variance, FPCA scores, persistent homology statistics, and graph theory measures (675 features). As discussed in Section 2, the regression parameters $\vtheta$ are estimated by maximizing the log-likelihood of the symmetric multinomial logistic regression subject to an elastic net penalty. The elastic net penalty is a mixture of ridge and lasso penalties and has two tuning parameters, $\lambda \ge 0$ a complexity parameter, and $0 \le \alpha \le 1$ a parameter balancing the ridge ($\alpha=0$) and lasso ($\alpha=1$) components. We choose values for these tuning parameters using cross-validation based on a nested cross-validation scheme similar to that proposed in Huttunen et al. (2011) that emphasizes the 50 labelled day two samples for obtaining error estimates. We consider a sequence of possible values for $\alpha$ lying in the set $\{0, 0.1, 0.2,\dots,1.0\}$ and fix $\alpha$ at one such value. With the given value of $\alpha$ fixed, we perform a 200-fold cross-validation. In each fold, the training data consists of all 677 samples from day one and a random sample of 25 of the 50 labelled day two samples. The remaining labelled day two samples are set aside as a validation set for the given fold. Within this fold, the 677+25 = 702 samples in the current training set are subjected to another 5-fold cross-validation over a sequence of $\lambda$ values to obtain an optimal $\lambda$ value for the given $\alpha$ \emph{and} training set. The resulting model is then used to classify the 25 validation samples resulting in a performance estimate $\epsilon_{\alpha,j}$ corresponding to the $j^{th}$ fold, $j=1,\dots,200$. The overall performance estimate for a given $\alpha$ is then obtained as the mean over the 200 folds $\epsilon_{\alpha} = \frac{1}{200}\sum_{j=1}^{200}\epsilon_{\alpha,j}$. This procedure is repeated for all $\alpha$ in $\left \{ 0,0.1,...,1.0 \right \}$. The optimal value for the tuning parameter $\alpha$ is that which corresponds to the smallest error $\epsilon_\alpha$.  Once the optimal $\alpha$ value has been determined, the optimal value for $\lambda$ is again chosen by 5-fold cross-validation as done previously, but now using all of the 727 training samples from both days. 

Table 1 lists the cross-validation predicted accuracy rates for each of the seven classifiers along with the test accuracy obtained from the 653 day two test samples. Had we participated in the competition, our choice of classifier would have been based on the cross-validation predicted accuracy rates. While all fairly close, the classifier incorporating detrended variance, FPC scores, and network features would have been chosen as our final model as this is one of two classifiers having the highest predicted accuracy rate 61.68\% and the fewest number of features of the two. The test accuracy from this classifier is 66.46\%, which is just short of 68.0\% obtained by the competition winners, but  higher than 63.2\% accuracy rate obtained by the first runner-up. \emph{Thus with our entry we would have finished in second place.} The confusion matrix for our classifier is presented in Table 2. Our classifier has highest accuracy for predicting the 'Chaplin' (92.8\%) video clips from the MEG data, and lowest accuracy for predicting the 'Football' (52.9\%) video clip.

\section{Discussion}
\label{sec:7}

We have reviewed the brain decoding problem in neuroscience and have discussed approaches from statistics, computational topology, and graph theory for constructing features for this high-dimensional classification problem. We have developed classifiers combining FPCA, persistent homology, and graph theoretic measures derived from mutual information networks. We have considered incorporating the features within a classifier based on symmetric multinomial logistic regression incorporating elastic net regularization and have applied our approach to a real brain decoding competition dataset illustrating good performance. Overall, examining the results in Table 1 we see that those classifiers incorporating FPC scores all perform quite well, with test accuracy scores being higher than predicted accuracy scores. It is not clear to us what aspect of the FPC scores allows for this increase and we are currently investigating this. Regarding the global features, there seems to be a small advantage gained in incorporating the network features but nothing gained by incorporating persistent homology. We emphasize that this is only for a single dataset and experimental paradigm. Performance on other brain decoding datasets may yield different results in particular as the samples considered in our application were based on fairly short 1-second recordings. We intend to continue investigating the utility of persistent homology and FPCA for decoding problems involving longer recordings and different experimental paradigms (involving face perception). Aside from classification, both techniques can also be used to explore and summarize novel aspects of neuroimaging data. Finally, given the interesting results we have observed with the classifiers incorporating FPCA, we are exploring the use of more general approaches based on nonlinear manifold representations for functional data such as those recently proposed by Chen and M\"{u}ller (2012).

\begin{acknowledgement}
This article is based on work from Nicole Croteau's MSc thesis. F.S. Nathoo is supported by an NSERC discovery grant and holds a Tier II Canada Research Chair in Biostatistics for Spatial and High-Dimensional Data.  
\end{acknowledgement}

%%%%%%%%%%%%%%%%%%%%%%%% referenc.tex %%%%%%%%%%%%%%%%%%%%%%%%%%%%%%
% sample references
% %
% Use this file as a template for your own input.
%
%%%%%%%%%%%%%%%%%%%%%%%% Springer-Verlag %%%%%%%%%%%%%%%%%%%%%%%%%%
%
% BibTeX users please use
% \bibliographystyle{}
% \bibliography{}
%

\begin{table}
\caption{Results from the brain decoding competition dataset. Baseline test accuracy is 23.0\% (chance level); competition winners achieved 68.0\% and second place was 63.2\%. Note that 'PH' refers to the 12 features derived using persistent homology.}
\label{tab:1}       % Give a unique label
%
% Follow this input for your own table layout
%
\begin{tabular}{lll}
\hline\noalign{\smallskip}
Classifier & CV Predicted Accuracy\,\,\,\, & Test Accuracy  \\
\noalign{\smallskip}\svhline\noalign{\smallskip}
Detrended variance & 60.90\% & 61.26\%\\
Detrended variance + FPCA  & 60.90\% & 65.54\% \\
Detrended variance + Network Features & 60.46\% & 61.41\%\\
Detrended variance + PH  & 60.44\% & 61.10\% \\
\bf{Detrended variance + FPCA  +  Network Features} & \bf{61.68\%} & \bf{66.46\%} \\
Detrended variance + FPCA  + PH & 60.72\% & 64.01\% \\
Detrended variance + FPCA  +  Network Features + PH\,\,\,\,\,\,\, & 61.68\% & 65.24\% \\
\noalign{\smallskip}\hline\noalign{\smallskip}
\end{tabular}
%$^a$ Table foot note (with superscript)
\end{table}

\begin{table}
\caption{Confusion matrix summarizing the performance on the test data for the classifier incorporating detrended variance, FPCA, and network features. }
\label{tab:1}       % Give a unique label
    \begin{tabular}{|l|lllll|}
%\hline\noalign{\smallskip}
    \hline
    Predicted Stimulus   $\backslash\backslash$     True Stimulus & Artificial \,\,\,& Nature\,\,\, & Football\,\,\, & Mr. Bean\,\,\, & Chaplin\,\,\, \\ \hline
    Artificial  & 90         & 27     & 28       & 6        &  3      \\
    Nature    & 39         &  98    & 16       & 6        & 0       \\
    Football  & 14         & 12     & 54       &  12      & 4       \\
    Mr. Bean  & 5          & 11     &  4       & 76       & 2       \\
    Chaplin   & 2          &   3    & 0        & 25       & 116     \\ \hline
    \end{tabular}
\end{table}

\begin{figure}
\centering
\begin{tabular}{c}
\includegraphics[height=0.5\textwidth]{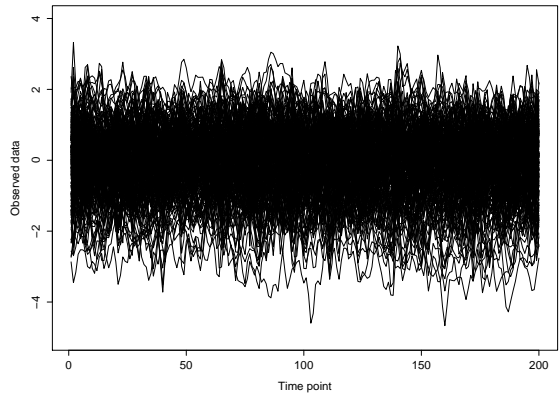} \\
(a)\\
%\hspace{-2.8em}
\includegraphics[height=0.4\textwidth]{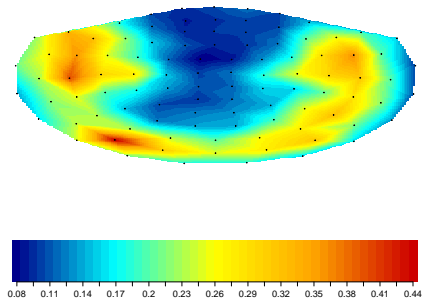} \\
(b)\\
\end{tabular}
\caption{A single sample from the training data: panel (a) - depicts the MEG (magnetic field) signals $Y_{li}(t)$ representing the evoked response collected at $n=204$ sensors; panel (b) - depicts the variance of the signal (after removal of linear trend) at $102$ locations. The map is a 2-dimensional projection of the sensor array with the black dots representing the sensor locations. There are 2 sensors at each location (each oriented differently) and the variance computed from each of the sensors is averaged to obtain a single value (for the purpose of visual summary only).}
\end{figure}

\begin{figure}
\includegraphics[height=.75\textwidth]{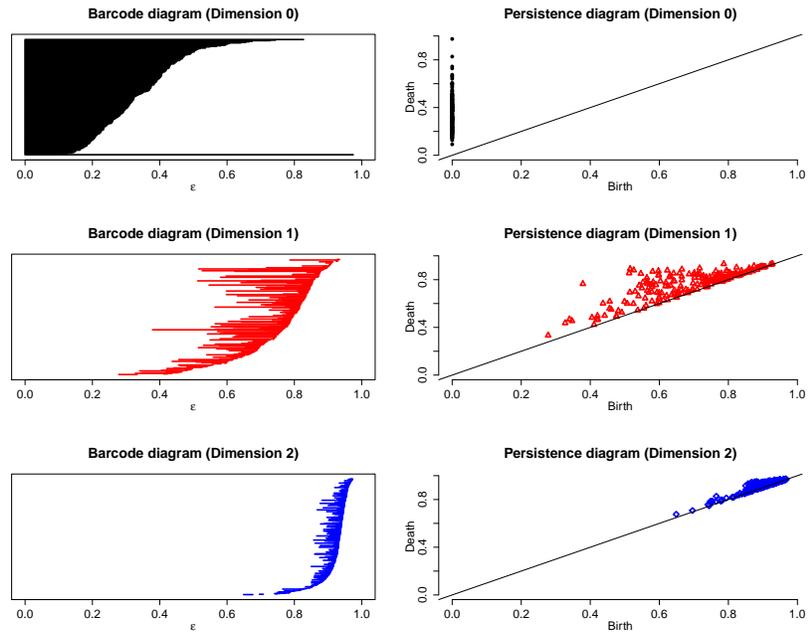} \\
\caption{Persistent homology computed for the single training sample depicted in Figure 1. The first column displays the barcodes for dimension $p = 0, 1, 2$ in each of the three rows respectively, and the second column displays the corresponding persistence diagrams.}
\label{comparisons}
\end{figure}

\begin{figure}
\centering
\includegraphics[height=0.75\textwidth]{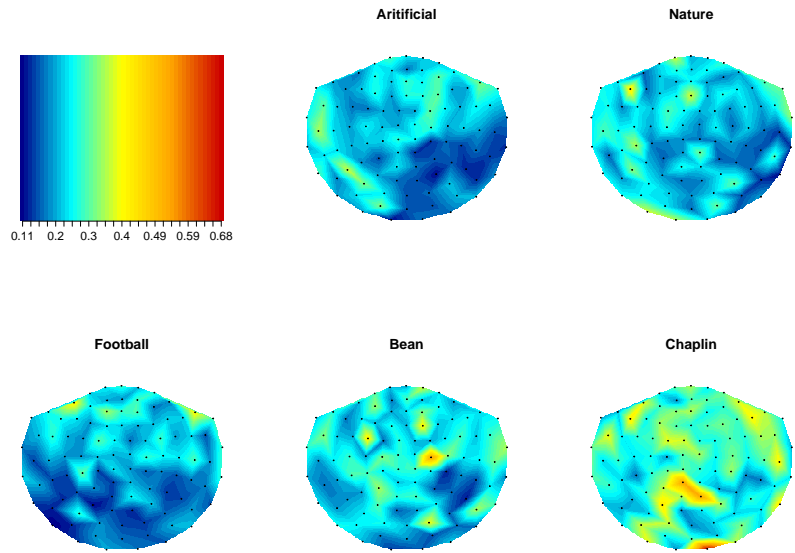} \\
\caption{Spatial variation of the detrended variance by stimulus class. Each map is a 2-dimensional projection of the sensor array with the black dots representing the sensors. At each sensor we fit a linear regression on time point and compute the variance of the residuals as the feature. There are 2 sensors (each oriented differently) at each of $102$ locations. For the purpose of visual summary, we average the two variance measures for each location and then further average across all training samples within a given stimulus class. We then map the resulting averaged measures across the scalp.}
\end{figure}

\begin{figure}
\centering
\begin{tabular}{cc}
\includegraphics[height=0.45\textwidth]{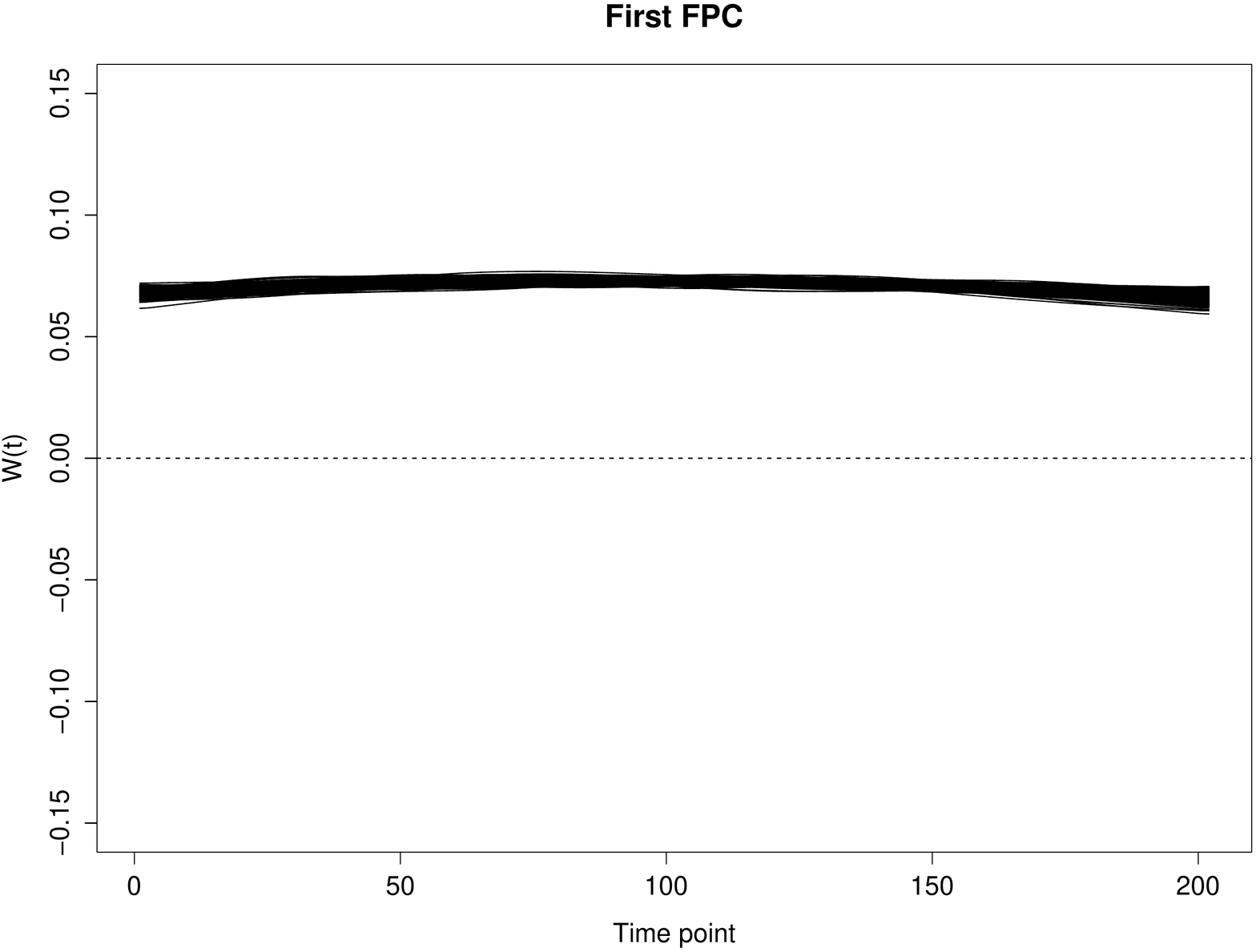} & \includegraphics[height=0.45\textwidth]{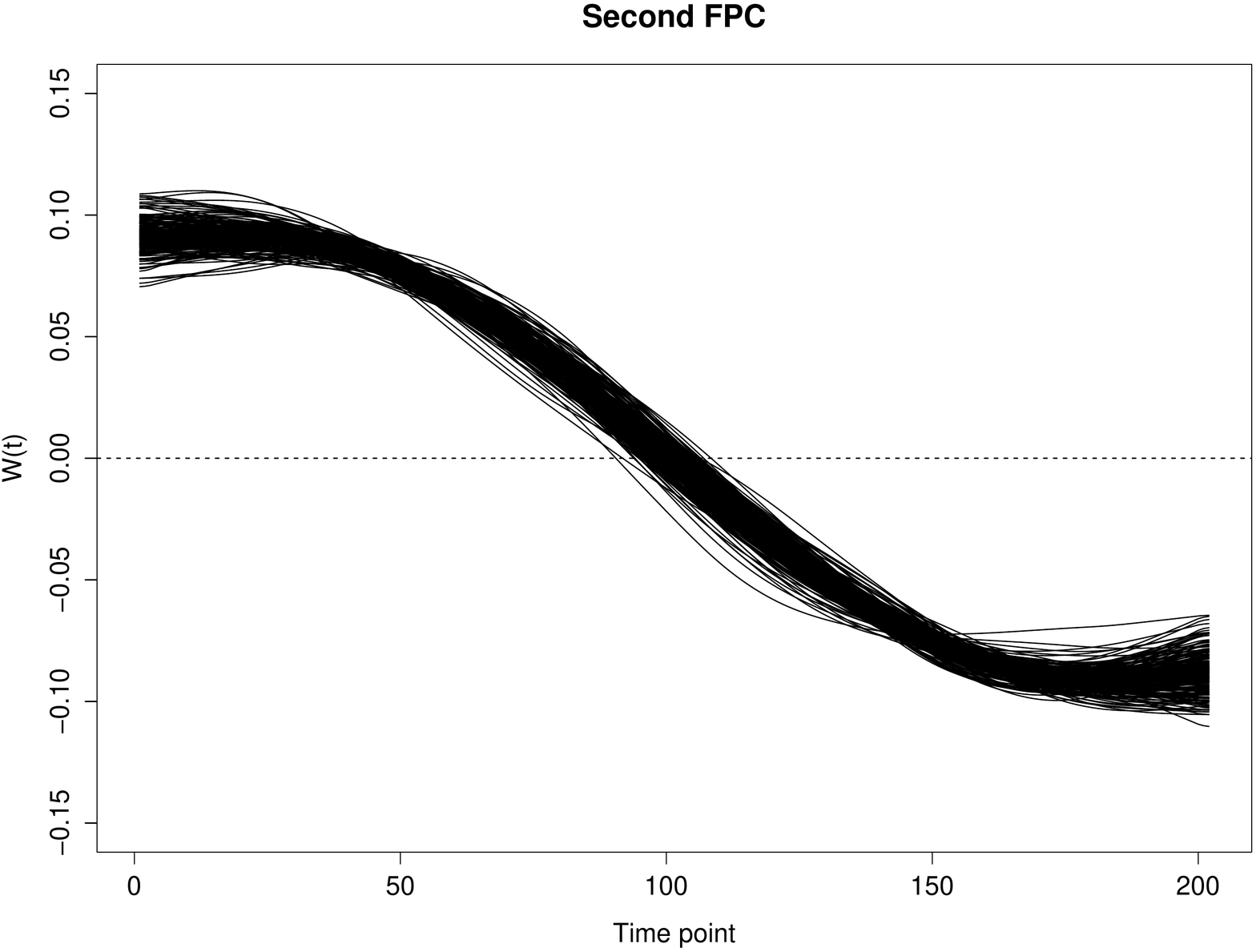} \\
(a) & (b)\\
%\hspace{-2.8em}
\includegraphics[height=0.45\textwidth]{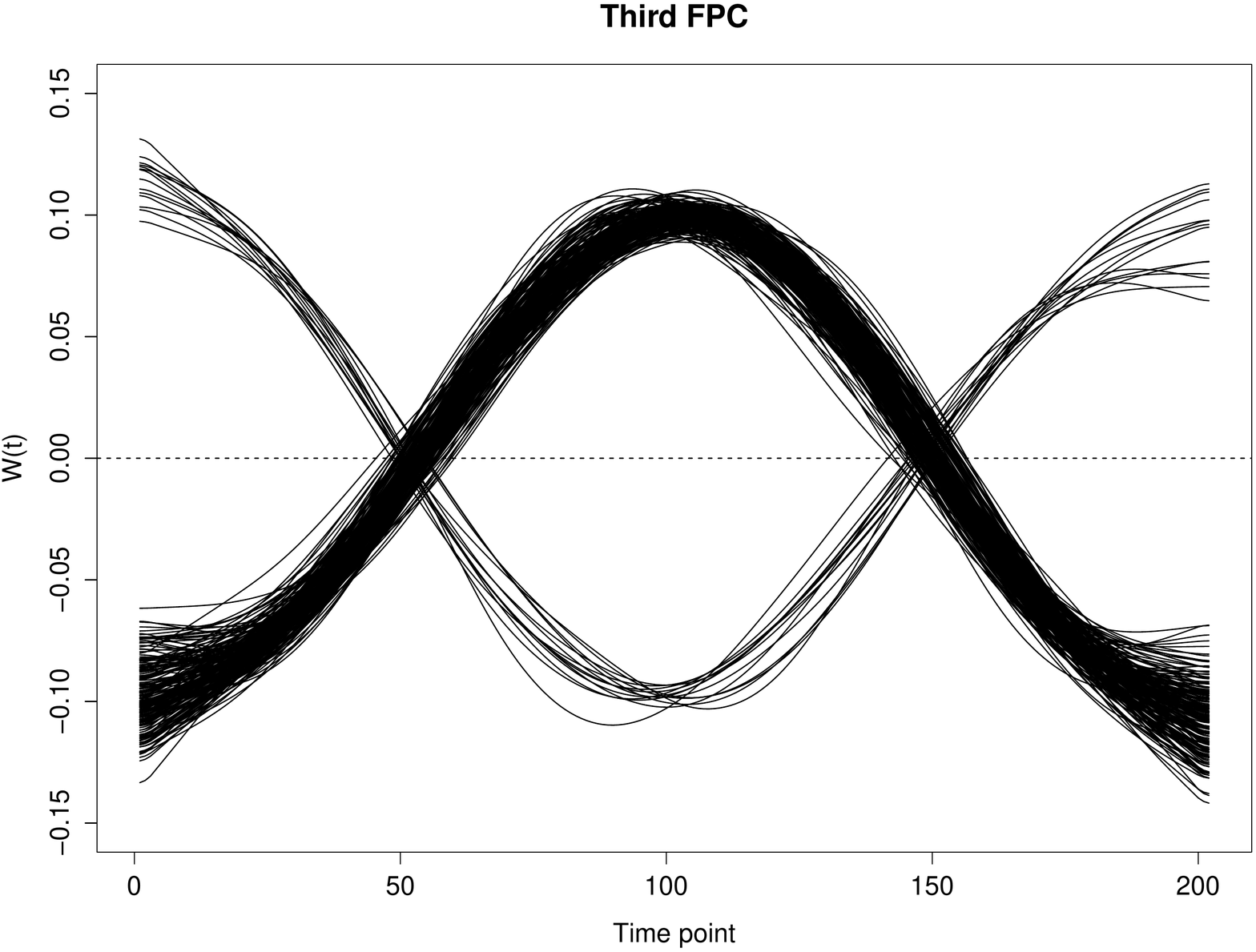} & \includegraphics[height=0.45\textwidth]{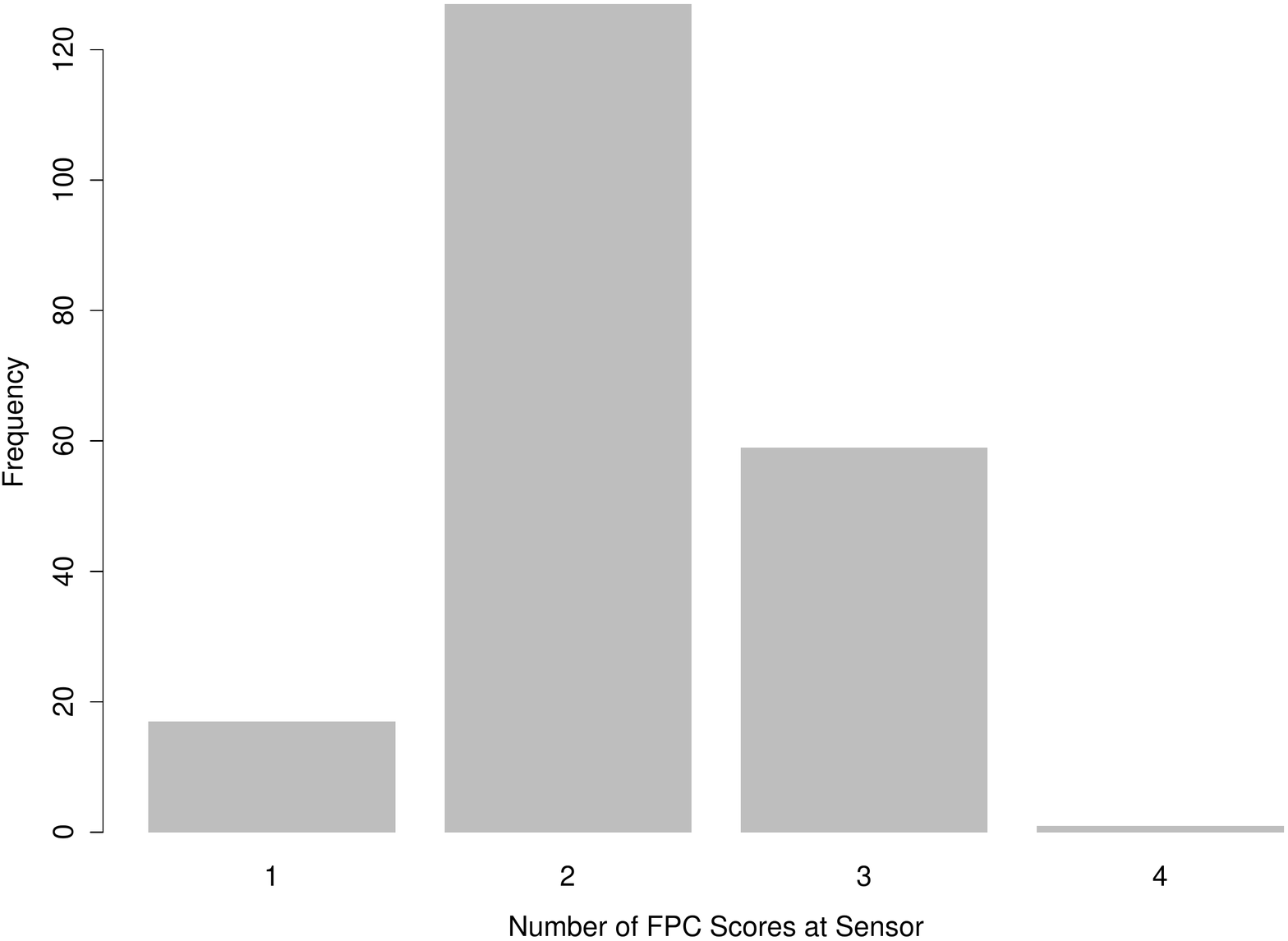} \\
(c) & (d)\\\\\\
\end{tabular}
\caption{FPCA applied to the training data: panel (a) - the first FPC at each sensor; panel (b) - the second FPC at each sensor; panel (c) - the third FPC at each sensor; (d) - the distribution of the smallest number of FPCs required to explain at least 90\% of the variance at each sensor.}
\label{comparisons}
\end{figure}

\begin{figure}
\centering
\includegraphics[height=0.75\textwidth]{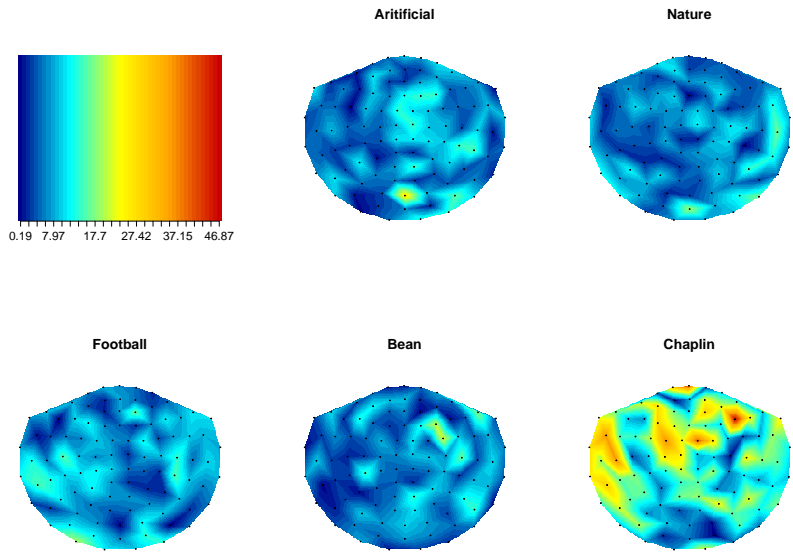} \\
\caption{Spatial variation of the first FPC score by stimulus class. Each map is a 2-dimensional projection of the sensor array with the black dots representing the sensor locations. There are 2 sensors (each oriented differently) at each of $102$ locations. For the purpose of visual summary, we average the absolute value of the 2 scores at each location and then further average across all training samples within a given stimulus class. We then map the resulting averaged measures across the scalp.}
\end{figure}

\end{document}